\documentclass[11pt,a4paper]{article}
\usepackage[utf8]{inputenc}
\usepackage[T1]{fontenc}
\usepackage{lmodern}
\usepackage[margin=2.5cm]{geometry}
\usepackage{graphicx}
\usepackage{booktabs}
\usepackage{array}
\usepackage{longtable}
\usepackage{multirow}
\usepackage{amsmath,amssymb}
\usepackage{hyperref}
\usepackage{xcolor}
\usepackage{enumitem}
\usepackage{float}
\usepackage[font=small,labelfont=bf]{caption}
\usepackage{authblk}
\usepackage{microtype}

\hypersetup{colorlinks=true,linkcolor=blue!60!black,citecolor=blue!60!black,urlcolor=blue!60!black}

\title{\textbf{UVLM: A Universal Vision-Language Model Loader\\for Reproducible Multimodal Benchmarking}}
\author[1]{Joan Perez\thanks{Corresponding author. mail: jperez@urbangeoanalytics.com ORCID: \href{https://orcid.org/0000-0003-3003-0895}{0000-0003-3003-0895}}}
\author[2]{Giovanni Fusco}
\affil[1]{Urban Geo Analytics, France}
\affil[2]{Universit\'e C\^ote d'Azur--CNRS--AMU--Avignon Universit\'e, ESPACE, France}
\date{}

\begin{document}
\maketitle

\begin{abstract}
Vision-Language Models (VLMs) have emerged as powerful tools for image understanding tasks, yet their practical deployment remains hindered by significant architectural heterogeneity across model families. This paper introduces UVLM (Universal Vision-Language Model Loader), a Google Colab-based framework that provides a unified interface for loading, configuring, and benchmarking multiple VLM architectures on custom image analysis tasks. UVLM currently supports two major model families --- LLaVA-NeXT and Qwen2.5-VL --- which differ fundamentally in their vision encoding, tokenization, and decoding strategies. The framework abstracts these differences behind a single inference function, enabling researchers to compare models using identical prompts and evaluation protocols. Key features include a multi-task prompt builder with support for four response types (numeric, category, boolean, text), a consensus validation mechanism based on majority voting across repeated inferences, a flexible token budget (up to 1,500 tokens) enabling users to design custom reasoning strategies through prompt engineering, and a built-in chain-of-thought reference mode for benchmarking. UVLM is designed for reproducibility, accessibility, and extensibility and as such is freely deployable on Google Colab using consumer-grade GPU resources. The paper also presents the first benchmarking of different VLMs on tasks of increasing reasoning complexity using a corpus of 120 street-view images.
\end{abstract}
\noindent\textbf{Keywords:} Vision-Language Models; Benchmarking; LLaVA; Qwen; Multimodal inference; Chain-of-thought; Consensus validation

\section{Motivation and significance}\label{sec:motivation}

Vision-Language Models (VLMs) represent a class of multimodal architectures that combine a visual encoder with a large language model decoder, enabling systems to interpret images and respond to natural language queries implying reasoning about their content. In recent years, the field has witnessed a rapid proliferation of VLM architectures, with prominent families including LLaVA \cite{ref1}, Qwen-VL \cite{ref2}, BLIP-2 \cite{ref3}, and InternVL \cite{ref4}, among others. Each family introduces its own design choices regarding vision encoding, cross-modal alignment, tokenization strategies, and decoding protocols. While this diversity has accelerated progress in multimodal AI, it has also created a significant practical challenge: researchers who wish to compare models across families must write and maintain separate inference pipelines for each architecture, even when the evaluation task is identical.

This fragmentation is particularly problematic for applied researchers outside the core machine learning community. In fields such as urban studies, remote sensing, medical imaging, or environmental monitoring, practitioners increasingly seek to leverage VLMs as zero-shot visual analysis tools. However, deploying even a single model requires understanding architecture-specific processor classes, chat template formats, image preprocessing pipelines, and generation configurations. Deploying multiple models for systematic comparison multiplies this complexity. As a result, many applied studies rely on a single model without benchmarking alternatives, limiting the robustness and generalizability of their findings.

The two VLM families currently supported by UVLM --- LLaVA-NeXT \cite{ref1,ref5} and Qwen2.5-VL \cite{ref2} --- illustrate the nature of this heterogeneity. LLaVA adopts a comparatively simple architecture: a CLIP-based vision encoder \cite{ref6} connected to a large language model through a trained MLP projection layer. The inference pipeline relies on a dedicated \texttt{LlavaNextProcessor} class that handles image-text interleaving, and the raw output is decoded directly from the full generated sequence, requiring post-processing to strip prompt tokens and instruction markers. Qwen2.5-VL, by contrast, employs a redesigned Vision Transformer (ViT) with RMSNorm, SwiGLU activations, and window-based attention, supporting native-resolution image inputs. Its inference pipeline uses a generic \texttt{AutoProcessor} with separate vision processing utilities (process\_vision\_info) and requires explicit trimming of input token IDs from the generated output before decoding. These differences extend to memory management --- Qwen models require additional visual token budget parameters (min/max pixel constraints) --- and to generation configuration, where Qwen loads settings from the model config object rather than accepting them as direct arguments. These are not superficial API variations; they reflect fundamentally different transformer logic in how visual information is encoded, merged with text tokens, and decoded into language.

UVLM was developed to address this challenge by providing a single, unified interface that abstracts away architecture-specific details while preserving full control over inference parameters. The tool originated in the context of the SAGAI project (Streetscape Analysis with Generative Artificial Intelligence) \cite{ref7}, a workflow for scoring street-level urban scenes using VLMs and open-access geospatial data, and the emc2 project \cite{ref8}, a European initiative studying the 15-minute city model in suburban areas, which has also produced open-source geospatial tools such as PPCA \cite{ref9}. Within these projects, researchers needed to systematically evaluate how different VLMs perform on identical visual analysis tasks --- such as estimating street frontage length, counting pedestrian entrances, and classifying building types from street-level photographs --- using exactly the same prompts and evaluation protocols. Writing and debugging separate pipelines for each model was both error-prone and time-consuming, motivating the creation of a unified loader.

UVLM is implemented as a Google Colab notebook and is designed to be freely accessible to any researcher with a Google account, without requiring local hardware or specialized infrastructure. It leverages the Hugging Face Transformers library \cite{ref10} and the BitsAndBytes quantization library \cite{ref11} to support models ranging from 3B to 110B parameters across multiple precision modes (FP16, 8-bit, 4-bit quantization). The framework is modular by design, and future releases will progressively integrate additional VLM families (e.g., InternVL \cite{ref4}, BLIP-2 \cite{ref3}, CogVLM) as their adoption grows in the research community.

\noindent The UVLM source code (v2.2.2) is publicly available under the Apache~2.0 license at \url{https://github.com/perezjoan/UVLM}.

\section{Software description}\label{sec:software}
\subsection{Software architecture}\label{sec:architecture}

UVLM is organized as a three-block sequential notebook, where each block handles a distinct stage of the VLM benchmarking workflow (Figure 1). This architecture ensures that the model loading phase (Block 1) is executed only once per session, while inference configuration (Block 2) and batch execution (Block 3) can be re-run independently as the user iterates on prompts or processes different image sets.

\begin{figure}[htbp]
\centering
\includegraphics[width=0.9\textwidth]{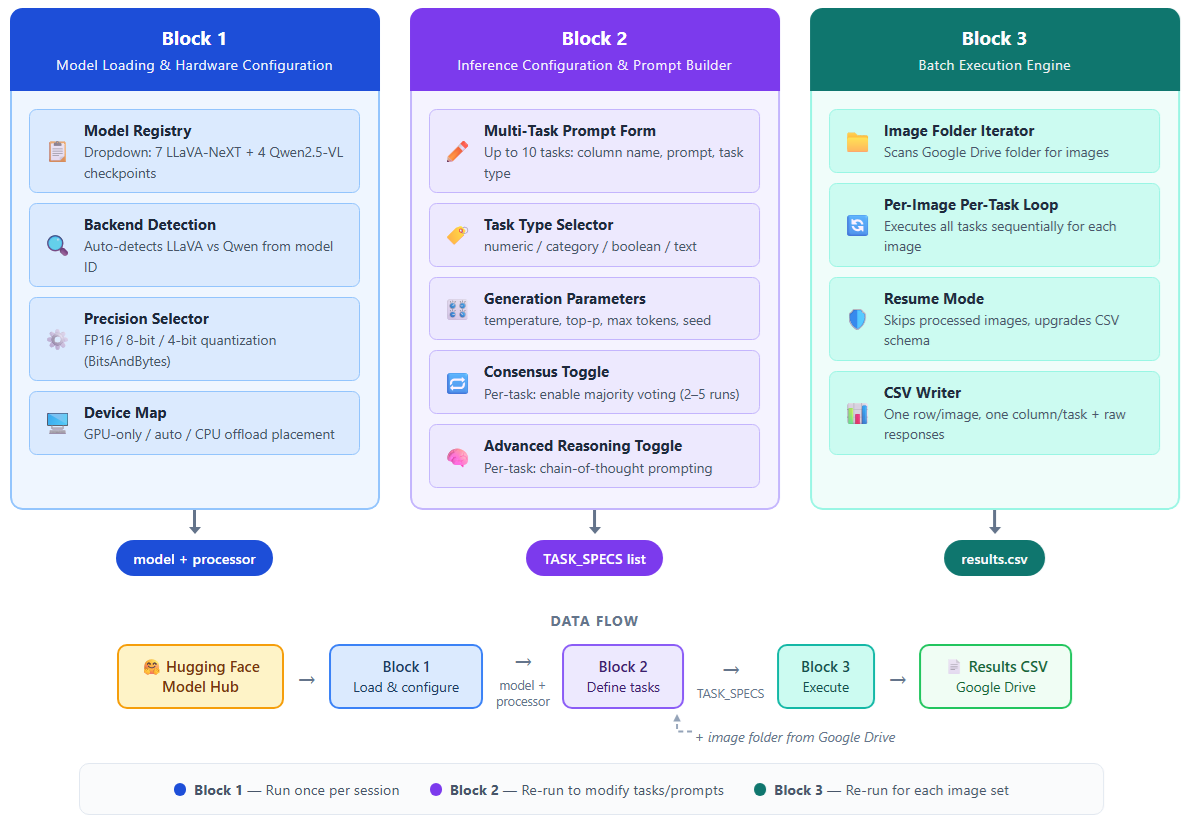}
\caption{UVLM software architecture. The notebook is organized into three sequential blocks. Block 1 loads the selected VLM and configures hardware settings, Block 2 defines the analysis tasks through an interactive prompt builder and Block 3 processes all images in a folder and writes results to CSV.}
\label{fig:architecture}
\end{figure}

Block 1: Model Loading and Hardware Configuration. The first block installs required dependencies (transformers, accelerate, bitsandbytes, \texttt{qwen-vl-utils}) and presents a dropdown interface for model selection. The model registry currently includes seven LLaVA-NeXT variants (Mistral 7B, Vicuna 7B/13B, 34B, LLaMA3 8B, 72B, 110B) and four Qwen2.5-VL variants (3B, 7B, 32B, 72B Instruct). Upon selection, the block automatically detects the model backend (LLaVA or Qwen) and loads the appropriate processor and model classes. Users can configure the precision mode (FP16, 8-bit, or 4-bit quantization via BitsAndBytes \cite{ref11}) and the device placement strategy (single GPU, automatic multi-device, or CPU offload). For Qwen models, additional visual token budget parameters (minimum and maximum pixel constraints) are configured to manage memory usage. The block outputs a loaded model-processor pair along with a persistent status widget displaying load time and device allocation.

Block 2: Inference Configuration and Prompt Builder. The second block provides a widget-based form for configuring up to ten analysis tasks. For each task, the user specifies a column name (for the output CSV), a text prompt, and a task type drawn from four categories: numeric (integer extraction), category (classification), boolean (yes/no), or text (free-form). Each task can optionally enable consensus validation (described in Section 2.4). The block also configures global generation parameters: temperature, top-p sampling, maximum token count (adjustable up to 1,500 tokens via a slider, enabling users to allocate sufficient generation budget for custom reasoning strategies), and an optional fixed random seed for reproducibility. A built-in advanced reasoning mode is also available as a reference implementation for chain-of-thought benchmarking (described in Section 2.3). Internally, Block 2 defines the core \texttt{run\_inference()} function, which implements the backend-agnostic forward pass, and the \texttt{parse\_response()} function, which applies type-specific output parsing.

Block 3: Batch Execution Engine. The third block iterates over all images in a user-specified Google Drive folder, executes all configured tasks sequentially for each image, and writes results to a CSV file (one row per image, one column per task). The engine supports resume mode, detecting already-processed images and skipping them to avoid redundant computation. It also implements schema upgrading: if the user adds new tasks between runs, the engine appends the missing columns to the existing CSV rather than overwriting it. Error handling is per-task --- if inference fails for one task on a given image, the remaining tasks are still executed, and the error is logged as "NA" in the corresponding cell. Additionally, UVLM includes built-in truncation detection: after each inference call, the exact number of generated tokens is compared against the token limit, and responses that hit the ceiling are flagged in per-task \texttt{\{column\}\_truncated} columns in the CSV with a console warning, allowing users to identify insufficient token budgets without post-hoc analysis.

\subsection{Dual-backend inference}\label{sec:dualbackend}

The central technical contribution of UVLM is its unified inference abstraction, which routes each call to the appropriate backend pipeline based on the loaded model family. The two pipelines differ substantially in how they process the image-text input and decode the model output.

LLaVA-NeXT pipeline. The LLaVA backend constructs a conversation object following the chat template format, with the image embedded as a typed content block alongside the text prompt. The \texttt{LlavaNextProcessor} applies the chat template, tokenizes the combined input, and returns tensors ready for generation. After the forward pass, the full output sequence --- including the echoed prompt --- is decoded, and the response is extracted by stripping architecture-specific markers (e.g., \texttt{[/INST]}, \texttt{ASSISTANT:} patterns) that vary depending on the base LLM (Mistral, Vicuna, LLaMA3).

Qwen2.5-VL pipeline. The Qwen backend constructs a message list with explicit image and text content types. Before tokenization, a dedicated \texttt{process\_vision\_info()} function extracts and preprocesses the visual inputs separately. The \texttt{AutoProcessor} then tokenizes text and visual features jointly. Generation uses a \texttt{GenerationConfig} object loaded from the model configuration. Crucially, the Qwen pipeline requires an explicit token-trimming step after generation: the prompt token IDs are subtracted from the output sequence before decoding, as the model returns the full concatenated sequence. This contrasts with the LLaVA approach, where prompt removal is performed on the decoded text via string matching.

\begin{figure}[htbp]
\centering
\includegraphics[width=\textwidth]{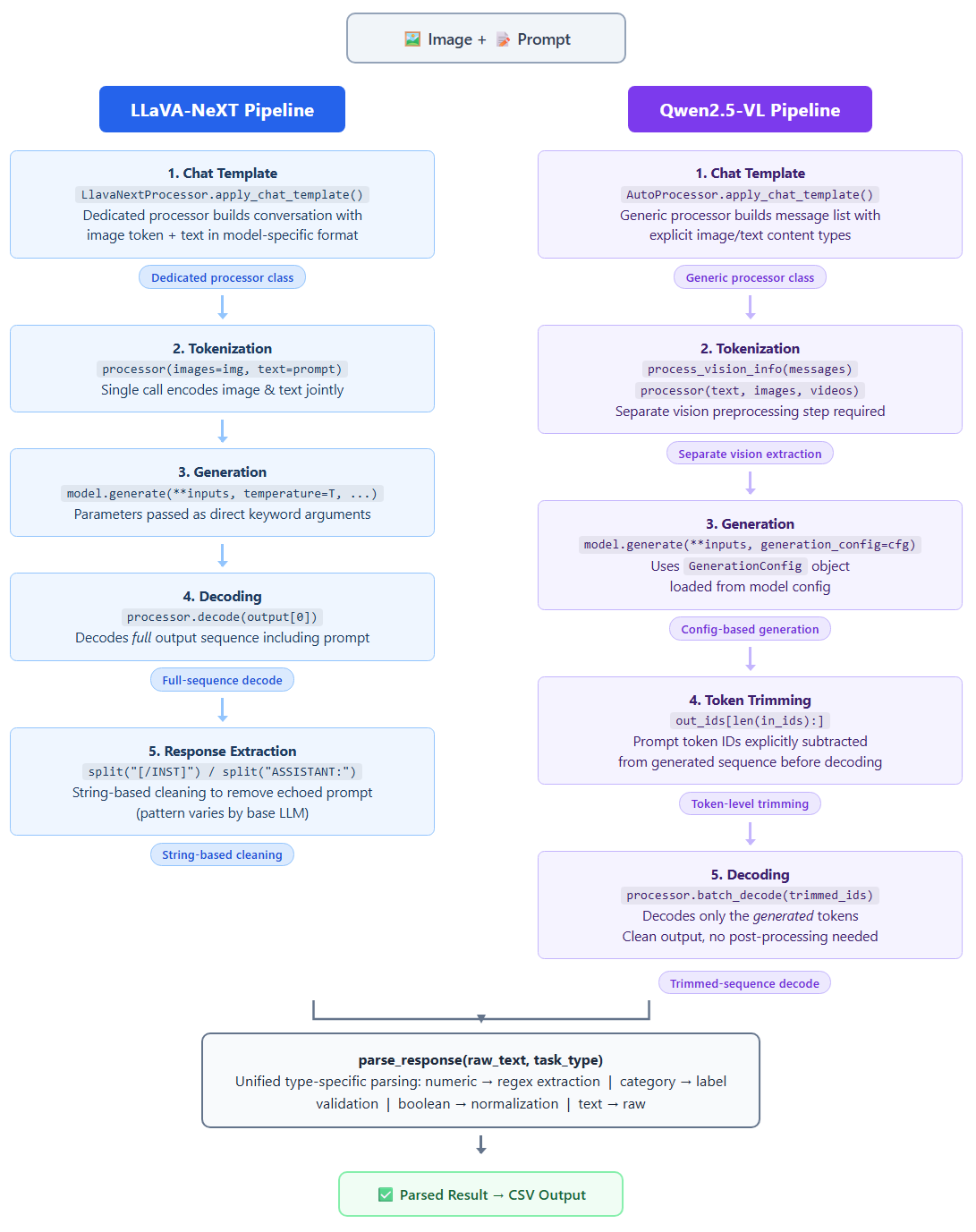}
\caption{Comparative diagram of the dual-backend inference pipelines in UVLM. Both pipelines receive the same image--prompt pair and converge at the unified response parser.}
\label{fig:pipelines}
\end{figure}

Both pipelines converge at the \texttt{parse\_response()} function, which applies task-type-specific parsing to the cleaned text output. For numeric tasks, the parser extracts the last number found in the response via regular expression --- this design choice ensures that when advanced reasoning produces intermediate numbers in its explanation, the final answer is captured. For category tasks, it strips common response prefixes (e.g., "The answer is:", "Based on the image,") and returns the cleaned label. For boolean tasks, it normalizes variations (yes/no/true/false) to a binary value. For text tasks, the raw response is returned as-is. When parsing fails, the value "NA" is recorded, and subsequent processing stages handle these missing values explicitly (see Section 2.4).

\subsection{Prompt engineering and reasoning support}\label{sec:prompts}

UVLM structures each task prompt as a concatenation of four elements: a role instruction (shared across all tasks, defining the model's persona), a task description (the specific question), a theory section (definitions, rules, and edge cases relevant to the task), and a format specification (expected output structure). This modular prompt architecture allows users to control prompt complexity independently of the underlying model, facilitating systematic experiments on prompt sensitivity.

Tasks requiring multi-step visual reasoning --- such as estimating distances by counting reference objects of known size --- benefit from chain-of-thought (CoT) prompting \cite{ref12}, where the model is encouraged to decompose complex analysis into explicit intermediate steps before producing a final answer. In UVLM, users can implement custom reasoning strategies by writing task prompts that request step-by-step explanations and increasing the maximum token budget via the slider (up to 1,500 tokens) to accommodate longer outputs. This approach gives full control over the reasoning structure and token allocation for each specific use case. As a convenience for benchmarking, UVLM also provides a built-in advanced reasoning reference mode. When enabled for a task, the format instruction is automatically replaced with a standardized CoT directive that asks the model to first describe its observations, then reason step by step, and finally provide a formatted answer on the last line (e.g., \texttt{ANSWER: <integer>} for numeric tasks). The token budget is automatically set to 1,024 tokens to accommodate the reasoning trace. The response parser then scans the last five lines of the output for the \texttt{ANSWER:} pattern and applies the standard type-specific parser to the extracted value. If the model fails to follow the expected format --- a common occurrence with smaller models --- the parser falls back to the standard extraction logic applied to the full response, ensuring graceful degradation rather than systematic failure.

The built-in reasoning mode serves primarily as a standardized reference for benchmarking: it applies the same CoT template across all models and tasks, enabling controlled comparisons of reasoning capability. In practice, users are encouraged to design their own reasoning prompts tailored to their specific tasks, using the max-token slider to provide adequate generation budget. Whether using custom prompts or the built-in mode, the reasoning trace is stored alongside the final answer in a dedicated CSV column \texttt{\{column\}\_reasoning}, allowing users to inspect and validate the model's reasoning process.

\subsection{Consensus validation}\label{sec:consensus}

VLM outputs are inherently stochastic when temperature sampling is used, and even with greedy decoding, small variations in input processing can lead to different results. To improve reliability, UVLM implements a consensus validation mechanism inspired by the self-consistency principle \cite{ref13}: each task is executed multiple times on the same image, and the final answer is determined by majority voting across all runs. Importantly, no single run is treated as primary --- all runs are equal peers in the voting process.

Consensus is enabled per task via a checkbox in the prompt builder form, with a configurable number of runs (2 to 5, default: 2). After all runs are completed, the \texttt{compute\_consensus()} function collects all parsed values, filters out any NA results (i.e., runs where the parser failed to extract a valid answer), and applies a majority vote over the remaining valid values. This filtering step ensures that parsing failures --- which can occur when a model does not follow the expected output format --- do not interfere with the voting process. For numeric tasks, an optional tolerance parameter allows grouping values within a specified percentage of each other as equivalent (0\% requires exact match). The output includes the consensus value, a boolean indicating whether consensus was reached (agreement ratio > 0.5), the agreement ratio itself (computed over all runs, including failed ones, to preserve the reliability metric), and the full list of individual run values. These are stored in additional CSV columns: \texttt{\{column\}\_consensus}, \texttt{\{column\}\_agreement}, and \texttt{\{column\}\_runs}.

\subsection{Supported models}\label{sec:models}

Table 1 lists all model checkpoints currently supported by UVLM. The internal registry is structured to facilitate maintenance and future expansion: models within an existing family are stored as dictionary entries, while new families require implementing the corresponding backend-specific sections of the inference function.

\begin{table}[htbp]
\centering
\caption{Supported VLM checkpoints in UVLM.}
\label{tab:models}
\small
\begin{tabular}{llcl}
\toprule
\textbf{Family} & \textbf{Model} & \textbf{Params} & \textbf{Checkpoint ID} \\
\midrule
LLaVA-NeXT & Mistral 7B & 7B & \texttt{llava-hf/llava-v1.6-mistral-7b-hf} \\
 & Vicuna 7B & 7B & \texttt{llava-hf/llava-v1.6-vicuna-7b-hf} \\
 & Vicuna 13B & 13B & \texttt{llava-hf/llava-v1.6-vicuna-13b-hf} \\
 & 34B & 34B & \texttt{llava-hf/llava-v1.6-34b-hf} \\
 & LLaMA3 8B & 8B & \texttt{llava-hf/llama3-llava-next-8b-hf} \\
 & 72B & 72B & \texttt{llava-hf/llava-next-72b-hf} \\
 & 110B & 110B & \texttt{llava-hf/llava-next-110b-hf} \\
\midrule
Qwen2.5-VL & 3B Instruct & 3B & \texttt{Qwen/Qwen2.5-VL-3B-Instruct} \\
 & 7B Instruct & 7B & \texttt{Qwen/Qwen2.5-VL-7B-Instruct} \\
 & 32B Instruct & 32B & \texttt{Qwen/Qwen2.5-VL-32B-Instruct} \\
 & 72B Instruct & 72B & \texttt{Qwen/Qwen2.5-VL-72B-Instruct} \\
\bottomrule
\end{tabular}
\end{table}

It should be noted that the largest checkpoints in each family --- particularly LLaVA-NeXT 72B and 110B, as well as Qwen2.5-VL 72B --- have memory requirements that exceed the capacity of a single consumer-grade GPU, even with 4-bit quantization. While UVLM includes these models in its registry and supports automatic device placement via \texttt{device\_map="auto"}, their effective use requires multi-GPU environments that go beyond the free-tier Google Colab offering. True multi-GPU parallelism with batched inference across devices is not yet implemented and is planned for a future release. In practice, models up to 34B parameters can be loaded on a single Colab GPU (T4 or A100) using 4-bit quantization, which covers the majority of the supported registry.

\section{Benchmark}\label{sec:benchmark}

To evaluate UVLM's capabilities and compare the supported model families, we conducted a systematic benchmark on a dataset of 120 street-level photographs of French urban frontages. This section describes the dataset, experimental protocol, evaluation metrics, and results. The complete benchmark report, including per-task per-model metrics, is provided as Supplementary Material 2.

\subsection{Dataset}\label{sec:dataset}

The benchmark dataset comprises 120 street-level images captured in the Nice--Côte d'Azur and Lille-Roubaix-Tourcoing metropolitan areas (France). The images depict a diverse range of street frontages, including dense urban blocks, suburban residential plots, and peri-urban areas with varying levels of vegetation. The image dataset is available and more specifically described on Zenodo \cite{ref14}.

A single expert annotator scored all 120 images on five analysis tasks, producing the human ground truth: (1) motor vehicle counting (integer): count of all motor vehicles visible in the image; (2) sidewalk detection (boolean): presence or absence of a sidewalk along the street frontage; (3) pedestrian entrance counting (integer): count of doors and gates giving pedestrian access from the street; (4) street frontage length (continuous, meters): estimated length of the street-facing boundary using reference objects of known size; and (5) vegetation type classification (ordinal, 1--6): a two-axis typology combining tree presence in the street space and vegetation density in abutting plots. Tasks were organized by increasing reasoning complexity. Task 1 involves simple object detection, a task traditionally handled by computer vision tools. Tasks 2--5 require reasoning about detected elements, including identifying relevant image sections and interpreting spatial relations. For instance, Task 3 requires counting pedestrian entrances only on plots abutting the relevant frontage segment (before the first side street). Double counting between setback gates and façade doors must be avoided. Car gates are excluded unless they are the sole access to a fenced plot and display indicators such as a doorbell, street number, or mailbox. The full prompt specifications for each task, including the shared role instruction, task-specific theory, and format directives, are provided as Supplementary Material 1.

\subsection{Experimental protocol}\label{sec:protocol}

We benchmarked eight model checkpoints from the two supported families, each in two inference modes (standard and advanced reasoning), yielding 16 model$\times$mode configurations. Table 2 summarizes the configurations and their hardware requirements.

\begin{table}[htbp]
\centering
\caption{Model configurations tested. GPU indicates the minimum requirement to avoid out-of-memory errors.}
\label{tab:configs}
\small
\begin{tabular}{llcc}
\toprule
\textbf{Model} & \textbf{GPU} & \textbf{Sec/img std} & \textbf{Sec/img reas} \\
\midrule
Qwen2.5-VL-3B & T4 & 0.73 & 17.23 \\
Qwen2.5-VL-7B & T4 & 2.17 & 16.23 \\
Qwen2.5-VL-32B & A100 & 1.06 & 60.50 \\
LLaVA Mistral 7B & T4 & 5.60 & 26.31 \\
LLaVA Vicuna 7B & T4 & 3.03 & 17.99 \\
LLaMA3 LLaVA 8B & L4 & 1.62 & 27.16 \\
LLaVA Vicuna 13B & A100 & 1.19 & 19.30 \\
LLaVA 34B & A100 & 5.36 & 160.01 \\
\bottomrule
\end{tabular}
\end{table}

The largest checkpoints in each family---Qwen2.5-VL-72B, LLaVA-NeXT 72B, and LLaVA-NeXT 110B---exceed single-GPU memory even with 4-bit quantization and were not included in this benchmark. Their effective use requires multi-GPU environments beyond the scope of UVLM's current single-GPU design.

In standard mode, each task used a maximum of 50 generated tokens with 3-run consensus validation. Since the prompt required a single numerical answer, the output was restricted to a single number. In advanced reasoning mode, the token budget was increased to 1024 to accommodate chain-of-thought reasoning, also with 3-run consensus validation. To reduce computational cost, the full chain-of-thought trace was saved only for the first run, even when the final answer differed across runs. All models used greedy decoding (temperature = 0) with a fixed random seed for reproducibility.

\begin{figure*}[htbp]

{\centering
\includegraphics[width=0.7\textwidth]{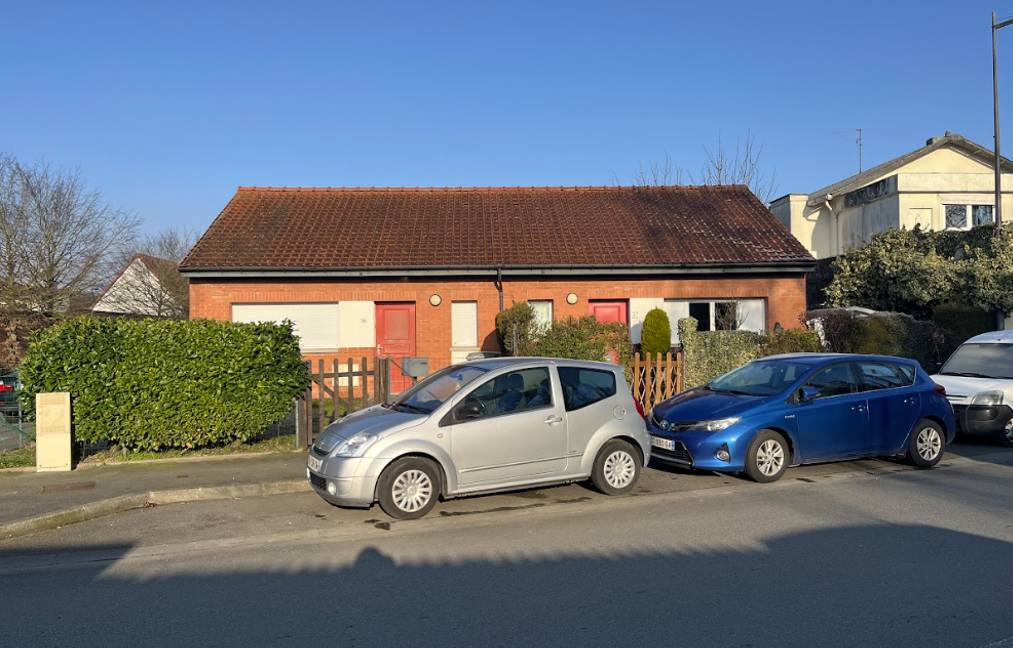}
\par}

\vspace{6pt}
\scriptsize
{\centering\textit{Visual input example, S27.jpg}\par}

\vspace{6pt}
\noindent\textbf{User: Task 1 --- motor vehicle counting (integer)}\\
\textbf{Human annotation:} 3\\
\textbf{Qwen2.5-VL-32B\_standard:} 3\\
\textbf{Qwen2.5-VL-32B\_reasoning:} There are three motor vehicles visible in the foreground of the image. The first vehicle is a silver car parked closest to the camera. It is fully visible and clearly identifiable as a car. The second vehicle is a blue car parked next to the silver car. It is also fully visible and identifiable as a car. The third vehicle is partially visible behind the blue car. Only the rear portion and part of the side are visible, but it appears to be another car based on its shape and size. There are no additional motor vehicles visible in the background. The focus is on the street frontage, and the background primarily consists of buildings, hedges, and a clear sky. The third vehicle is partially obscured by the blue car, but enough of its structure (e.g., rear lights, wheels, and body shape) is visible to confirm it is a motor vehicle. Bicycles and scooters are not present in the image, so they do not need to be counted. No other types of motor vehicles (e.g., vans, trucks, buses, motorcycles) are visible in the image. - Silver car: 1 - Blue car: 1 - Partially hidden car: 1 \#\#\# Answer: 3

\vspace{6pt}
\noindent\textbf{User: Task 2 --- sidewalk detection (boolean)}\\
\textbf{Human annotation:} 1\\
\textbf{Qwen2.5-VL-32B\_standard:} 1\\
\textbf{Qwen2.5-VL-32B\_reasoning:} The street frontage is the boundary between the public street space and the adjoining plots. In this image, the street frontage is defined by the buildings, fences, and landscaping along the left side of the street. The sidewalk extends along the entire visible length of the street frontage in the image. It starts from the left edge of the image and continues to the right, following the boundary of the properties. While there are some immaterial elements like the edge of the garden or the setback area, the presence of the sidewalk itself is a clear material element that delimits the street frontage. There is no visible perpendicular intersection in the image. However, the sidewalk continues along the entire visible street frontage without interruption. The sidewalk is present and runs along the entire visible street frontage, delimiting the boundary between the public street space and the adjoining plots. \#\#\# Conclusion: 1

\vspace{6pt}
\noindent\textbf{User: Task 4 --- street frontage length (continuous, meters)}\\
\textbf{Human annotation:} 15\\
\textbf{Qwen2.5-VL-32B\_standard:} 20\\
\textbf{Qwen2.5-VL-32B\_reasoning:} The street frontage is the boundary visible along the street, extending from the left edge of the image to the right edge. It includes material elements like buildings, fences, hedges, and immaterial elements like the edge of the garden or parking lot. The street frontage is bounded by the red-brick house on the left and extends to the white building on the right. The red-brick house has a clear facade with doors and windows. A hedge runs along the left side of the house, and there is a parked car in front of it. Two cars are parked in front of the red-brick house. These cars can serve as reference objects for estimating length. The street frontage continues beyond the red-brick house, including a white building and some greenery. The boundary appears to extend to the right edge of the image. Cars are a suitable reference object because they are clearly visible and have a known average length of 4 to 4.5 meters. Using cars will help estimate the length of the street frontage accurately. Car 1 (Silver): This car is fully visible and aligned parallel to the street. Its length is approximately 4 meters. Car 2 (Blue): This car is also fully visible and aligned parallel to the street. Its length is approximately 4 meters. Between the two cars and beyond the blue car, there is additional street frontage. By comparing the space to the length of the cars, we can estimate the remaining length. The space between the two cars appears to be about the width of one car (4 meters). Beyond the blue car, the street frontage extends further. This additional length appears to be roughly equivalent to another car's length (4 meters). Car 1: 4 meters. Space between cars: 4 meters. Car 2: 4 meters. Additional space beyond Car 2: 4 meters. Total estimated length = 4 + 4 + 4 + 4 = 16 meters. Objects farther from the camera appear shorter due to perspective. However, since the cars are relatively close to the camera and spread across the image, the perspective effect is minimal. The estimates remain reasonably accurate. Final Answer: 16

\caption{Qwen2.5-VL-32B outputs (standard and reasoning) as compared to human annotations for task 1, 2 and 4.}
\label{fig:example}
\end{figure*}

Figure 3 illustrates typical differences between the two inference modes. In standard mode, the model directly outputs a single numerical answer as required by the prompt, while advanced reasoning mode generates an explicit chain-of-thought before producing the final value. 

\subsection{Evaluation metrics}\label{sec:metrics}

We report task-appropriate metrics for each task type, plus a unified proximity score used for cross-task and cross-model ranking.

For sidewalk detection (boolean), we report accuracy, sensitivity, specificity, and Cohen's $\kappa$. For counting tasks (vehicles, entrances), we report mean absolute error (MAE), bias (signed mean error), exact match rate, $\pm$1 tolerance rate, and Pearson correlation. For length estimation, we additionally report mean absolute percentage error (MAPE) and $\pm$10 m tolerance rates. For vegetation classification (ordinal), we report exact match, $\pm$1 class tolerance, MAE in class units, and linearly weighted Cohen's $\kappa$.

We also report two reliability indicators: the NA rate (fraction of consensus runs where the parser failed to extract a valid answer) and the truncation rate (fraction of tasks where the response hit the token limit).

The \textbf{proximity score} is defined as:
\begin{equation}
\text{Proximity} = \frac{1}{N} \sum_{i=1}^{N} \max\!\left(0,\; 1 - \frac{|y_i - \hat{y}_i|}{R}\right)
\label{eq:proximity}
\end{equation}
where $y_i$ is the human value, $\hat{y}_i$ is the model prediction, and $R$ is the observed range of human values. Unlike exact matches, proximity rewards close answers: if the human scores 5 vehicles and the model predicts 6, the proximity score is approximately 87\%, not 0\%. The overall proximity is the unweighted mean across all five tasks and serves as the primary ranking metric.

\subsection{Results}\label{sec:results}

\begin{table}[htbp]
\centering
\caption{Overall model ranking by proximity score (\%). Higher is better. $^\dagger$~Reduced sample due to CUDA crashes on T4 (see Section~\ref{sec:limitations}). $^\ddagger$~Partial benchmark due to prohibitive computation time.}
\label{tab:ranking}
\small
\begin{tabular}{clccccccc}
\toprule
\textbf{\#} & \textbf{Model} & \textbf{N} & \textbf{T1 Veh.} & \textbf{T2 Side.} & \textbf{T3 Entr.} & \textbf{T4 Len.} & \textbf{T5 Veg.} & \textbf{Mean} \\
\midrule
1 & Qwen 32B reas & 120 & 97.7 & 85.0 & 90.4 & 82.5 & 84.2 & \textbf{88.0} \\
2 & Qwen 32B std & 120 & 98.1 & 80.8 & 92.3 & 55.4 & 89.8 & 83.3 \\
3 & LLaVA Vic7B std & 120 & 95.4 & 84.2 & 85.7 & 76.5 & 74.0 & 83.1 \\
4 & LLaVA Vic13B std & 120 & 96.1 & 80.8 & 86.2 & 56.0 & 70.3 & 77.9 \\
5 & Qwen 7B reas & 120 & 95.8 & 62.5 & 84.9 & 73.3 & 70.7 & 77.4 \\
6 & LLaVA Vic13B reas & 120 & 93.6 & 65.8 & 80.7 & 70.4 & 74.5 & 77.0 \\
7 & LLaVA Vic7B reas & 120 & 93.0 & 77.5 & 80.9 & 67.1 & 66.3 & 77.0 \\
8 & Qwen 3B reas & 118$^\dagger$ & 96.1 & 56.8 & 85.5 & 66.5 & 75.4 & 76.1 \\
9 & Qwen 3B std & 60$^\dagger$ & 96.5 & 75.0 & 77.5 & 70.1 & 60.7 & 76.0 \\
10 & Qwen 7B std & 120 & 97.5 & 40.0 & 89.0 & 70.1 & 80.7 & 75.4 \\
11 & LLaVA Mis7B std & 120 & 95.8 & 45.0 & 85.7 & 75.4 & 70.3 & 74.4 \\
12 & LLaVA 34B reas & 25$^\ddagger$ & 88.0 & 80.0 & 76.5 & 70.4 & 56.0 & 74.2 \\
13 & LLaVA Mis7B reas & 120 & 94.0 & 70.0 & 83.9 & 65.9 & 52.8 & 73.3 \\
14 & LLaMA3 8B reas & 120 & 93.7 & 46.7 & 84.9 & 66.3 & 65.3 & 71.4 \\
15 & LLaMA3 8B std & 120 & 95.9 & 51.7 & 86.0 & 32.8 & 60.3 & 65.3 \\
16 & LLaVA 34B std & 120 & 95.8 & 15.8 & 85.8 & 47.2 & 66.5 & 62.2 \\
\bottomrule
\end{tabular}
\end{table}

The full benchmark report is available as Supplementary Material 2. We will focus here on our main findings.

\paragraph{Best overall performance.}
Qwen2.5-VL-32B with advanced reasoning achieved the highest overall proximity (88.0\%), outperforming all other configurations. On motor vehicle counting, it reached an MAE of 0.25 with 81.7\% exact match and 95.0\% within $\pm$1. On sidewalk detection, accuracy was 85.0\%. On entrance counting, the MAE was 0.77 with 84.2\% within $\pm$1. On frontage length, the MAE was 5.8 m with 85.0\% of estimates within $\pm$10 m and a Pearson correlation of 0.48. On vegetation classification, exact match was 55.8\% with a linearly weighted $\kappa$ of 0.54 (moderate agreement).

\paragraph{Model size vs.\ performance.}
Larger models did not systematically outperform smaller ones. LLaVA Vicuna 7B in standard mode (83.1\%) ranked third overall, ahead of all LLaVA models up to 34B. LLaVA 34B in standard mode ranked last among all 16 configurations (62.2\%), with particularly poor sidewalk detection (15.8\% accuracy) and a length MAE of 17.6 m. This suggests that architectural fit and prompt compatibility matter more than raw parameter count.

\paragraph{Effect of advanced reasoning.}
Advanced reasoning improved performance for Qwen models: Qwen 32B gained +4.7 percentage points in overall proximity, driven primarily by a +19.5 point improvement on length estimation. Qwen 7B gained +2.0 points. For LLaVA models, the effect was mixed to negative: LLaMA3 LLaVA 8B gained +6.1 points overall, but most LLaVA-NeXT models showed degradation, particularly on counting tasks where verbose reasoning introduced parsing failures. The NA rate (parsing failures) was 10.9\% for LLaMA3 8B in reasoning mode versus 0.3\% in standard mode, indicating that the model frequently failed to produce the expected \texttt{ANSWER:} line.

\paragraph{Reliability.}
Qwen models exhibited near-perfect parsing reliability: zero NA and zero truncation in standard mode across all three model sizes. In reasoning mode, Qwen 32B achieved zero truncation on all 120 images within the 1024-token budget. By contrast, LLaMA3 LLaVA 8B in reasoning mode had a truncation rate of 16.8\%, meaning its chain-of-thought output was forcibly cut off on 101 out of 600 task$\times$image checks, which in turn caused the high NA rate.

\paragraph{Task difficulty.}
Across all models, motor vehicle counting was the easiest task, with proximity scores tightly clustered between 93.0\% and 98.1\% (a spread of only 5.1 percentage points). Vegetation classification was the hardest, with the best model reaching only 89.8\% proximity and a linearly weighted $\kappa$ of 0.54 (moderate agreement). Frontage length estimation showed the widest performance spread: proximity ranged from 32.8\% (LLaMA3 8B standard) to 82.5\% (Qwen 32B reasoning), a gap of 49.7 percentage points, reflecting the inherent difficulty of metric distance estimation from single images and the sensitivity of this task to both model capability and inference mode.

\paragraph{Computation cost.}
Computation time varied substantially across configurations (Table 2). In standard mode, Qwen 7B on a T4 processed each image in 2.17 s, while LLaVA 34B on an A100 required 5.36 s. The gap widened dramatically in reasoning mode: Qwen 7B remained efficient at 16.23 s/image, whereas LLaVA 34B required 160 s/image---nearly 5.3 hours for 120 images on an A100. This prohibitive computation time is why LLaVA 34B reasoning was limited to 25 randomly sampled images. These timings highlight that models requiring A100 GPUs impose significantly higher resource demands, making smaller models on consumer-grade GPUs (T4, L4) attractive for large-scale deployment.

\subsection{Practical limitations encountered during benchmarking}\label{sec:limitations}

Two practical issues arose during the benchmark. First, Qwen2.5-VL-3B running on a T4 GPU experienced intermittent CUDA device-side assertion failures during both standard and advanced reasoning modes. The T4 supports only FP16, whereas Qwen models are trained in BF16; the narrower dynamic range of FP16 can cause numerical overflow during generation, producing invalid tensor indices that crash the GPU context irreversibly. Two images were lost in reasoning mode, and half the standard-mode run was also affected (60 out of 120 images). Running Qwen2.5-VL-3B on a GPU with native BF16 support (e.g., A100 or L4) may resolve this issue, though this was not tested in the present benchmark.

Second, LLaVA 34B required approximately 160 seconds per image in reasoning mode on an A100 (5 tasks $\times$ 3 consensus runs $\times$ 1024 max tokens). The full 120-image benchmark was therefore limited to standard mode; only 25 randomly sampled images were processed in reasoning mode to provide a partial estimate of reasoning performance.

\section{Impact}\label{sec:impact}

UVLM is a fully open-source framework that addresses a practical gap in the current VLM ecosystem: the absence of a lightweight, accessible tool for systematically comparing vision-language models across architectures using identical evaluation protocols. By abstracting the substantial implementation differences between model families behind a unified interface, UVLM enables reproducible benchmarking that would otherwise require significant engineering effort.

While UVLM was developed in the context of urban research --- specifically within the SAGAI \cite{ref7} and emc2 \cite{ref8} projects for streetscape analysis --- its design is domain-agnostic. The four supported task types (numeric, category, boolean, text) and the free-form prompt builder accommodate a wide range of image analysis applications. Potential use cases include, but are not limited to: medical image interpretation (e.g., classifying radiological findings, counting anatomical structures), remote sensing (e.g., land use classification from satellite imagery, estimating building counts), environmental monitoring (e.g., detecting pollution indicators, classifying vegetation), agricultural assessment (e.g., crop health scoring, pest detection), infrastructure inspection (e.g., damage assessment, defect counting), and document analysis (e.g., extracting structured data from scanned forms). In each of these domains, the ability to benchmark multiple VLM architectures on the same task with the same prompts provides valuable evidence for model selection decisions.

The consensus validation and reasoning support features further broaden the tool's applicability. Consensus validation addresses the stochastic nature of VLM outputs, which is particularly important in applications where reliability matters --- for instance, when VLM predictions feed into downstream quantitative analyses or policy recommendations. The flexible token budget (up to 1,500 tokens) and modular prompt architecture allow users to design custom chain-of-thought strategies tailored to their specific tasks, while the built-in advanced reasoning mode provides a standardized reference for benchmarking multi-step visual inference.

The benchmark results (Section 3) demonstrate that UVLM successfully enables systematic comparison across model families and inference modes. The proximity score, which rewards close answers rather than requiring exact matches, revealed that Qwen2.5-VL-32B with advanced reasoning achieves the highest overall proximity score (88.0\%), while LLaVA Vicuna 7B in standard mode provides a competitive alternative (83.1\%) at a fraction of the computation cost. These findings also highlight that model size alone does not determine performance: LLaVA 34B ranked last among all configurations, underscoring the importance of prompt-model compatibility. Finally, UVLM's deployment on Google Colab with support for 4-bit quantization democratizes access to large VLMs. Researchers without access to dedicated GPU clusters can evaluate models with up to 7B parameters using free-tier Colab resources (T4 GPU), lowering the barrier to entry for multimodal AI research across disciplines.

\section{Conclusions}\label{sec:conclusions}

This paper presented UVLM, a universal vision-language model loader designed for reproducible multimodal benchmarking. The framework addresses the practical challenge of architectural heterogeneity across VLM families by providing a unified interface for model loading, inference, and evaluation. The current implementation supports two major families --- LLaVA-NeXT and Qwen2.5-VL --- which differ fundamentally in their vision encoding, tokenization, and decoding strategies, and abstracts these differences behind a single inference function.

The benchmark on 120 French street-level images, covering five urban analysis tasks across eight models in two inference modes, yielded several findings. First, Qwen2.5-VL-32B with advanced reasoning achieved the best overall performance (88.0\% proximity), with motor vehicle count reaching 95.0\% within $\pm$1 of the human value. Second, advanced reasoning improved spatial estimation tasks (frontage length) by up to 19.5 percentage points for Qwen models, but degraded performance for several LLaVA models due to parsing failures. Third, model size did not predict performance: the 7B Vicuna-based LLaVA (83.1\%) outperformed the 34B model (62.2\%). Fourth, Qwen models exhibited near-perfect parsing reliability (zero failures in standard mode), while LLaVA models in reasoning mode suffered NA rates up to 10.9\%. Overall, the diversity of the five benchmarking tasks suggests that the best-performing models (which in turn depend on prompting) reach proximity scores compatible with high-quality strategic streetscape assessments at spatial scales (e.g., several thousand streetscapes) that exceed feasible human analysis.

Several directions for future work are envisioned. First, support for additional VLM families --- including InternVL \cite{ref4}, BLIP-2 \cite{ref3}, CogVLM, DeepSeek-VL, Molmo, and GLM-V --- will be integrated to broaden the comparative scope of the framework. Second, multi-GPU batching across images (rather than the current sequential processing) will improve throughput for large-scale evaluations. Third, video frame analysis support will extend UVLM's applicability to temporal visual tasks. Fourth, an API mode for cloud deployment will enable integration with automated pipelines beyond the notebook interface. Finally, automatic prompt optimization techniques could further reduce the burden on users who lack expertise in prompt engineering. On the application side, UVLM will be integrated into the SAGAI workflow \cite{ref7} as its vision-language inference engine, replacing the current single-model implementation and enabling SAGAI users to benchmark and select the best-performing VLM for their specific streetscape analysis tasks.

By providing an accessible, reproducible, and extensible benchmarking framework, UVLM aims to support more rigorous and transparent evaluation practices in the rapidly evolving field of vision-language models, while enabling applied researchers across disciplines to leverage these models effectively for their specific tasks.

\clearpage
\appendix

\section*{Supplementary Material 1: UVLM Benchmark Prompts}
\addcontentsline{toc}{section}{Supplementary Material 1}

Each task is sent as a single prompt: ROLE + TASK + THEORY + FORMAT concatenated. The ROLE is shared across all tasks. Important: When advanced reasoning is enabled, the FORMAT field is automatically replaced by the code with the chain-of-thought template.

\paragraph{Shared ROLE}
You analyze a street-level image. Be precise and follow instructions exactly. The street frontage is the street-facing boundary between the public street space and the adjoining plots, on one side of the street only. It may consist of material elements (building facade, fence, wall, hedge) or immaterial elements (edge of a garden, parking lot, vacant plot, or setback area when no physical boundary exists).

\vspace{10pt}
\noindent\textbf{Task 1 --- Motor Vehicle Counting}
{\small
\begin{center}
\begin{tabular}{p{3cm}p{11cm}}
\midrule
Type / Column & task\_type: numeric    |    CSV column: vehicles \\
\midrule
ROLE & (shared ROLE above) \\
\midrule
TASK & Count the motor vehicles in the image. \\
\midrule
THEORY & All motor vehicles must be counted: cars, vans, trucks, buses, streetcars, motorcycles, etc. Do not count bicycles or scooters, even if they have an electric motor. Count all motor vehicles that are visually distinguishable. Motor vehicles partially hidden by other objects should still be considered if clear visual cues are present. \\
\midrule
FORMAT (std) & Answer with only one integer number, nothing else. \\
\midrule
FORMAT (reas) & (auto-generated by code --- do not fill in) \\
\bottomrule
\end{tabular}
\end{center}
}

\vspace{10pt}
\noindent\textbf{Task 2 --- Sidewalk Detection }
{\small
\begin{center}
\begin{tabular}{p{3cm}p{11cm}}
\midrule
Type / Column & task\_type: boolean    |    CSV column: sidewalk \\
\midrule
ROLE & (shared ROLE above) \\
\midrule
TASK & Detect the presence of a sidewalk delimiting the street frontage visible in the image. \\
\midrule
THEORY & The street frontage extends to the first clearly visible perpendicular intersection. If none is visible, stop at the last visible point of the street-facing boundary. Consider as sidewalk both a raised pavement and a protected pedestrian path bordering the street. If the sidewalk does not cover the entire street frontage, still consider it as present. \\
\midrule
FORMAT (std) & Answer with only one word: yes or no \\
\midrule
FORMAT (reas) & (auto-generated by code --- do not fill in) \\
\bottomrule
\end{tabular}
\end{center}
}

\vspace{180pt}
\noindent\textbf{Task 3 --- Pedestrian Entrance Counting}
{\small
\begin{center}
\begin{tabular}{p{3cm}p{11cm}}
\midrule
Type / Column & task\_type: numeric    |    CSV column: entrances \\
\midrule
ROLE & (shared ROLE above) \\
\midrule
TASK & Count all pedestrian entrances to the plots abutting the street frontage visible in the image. \\
\midrule
THEORY & The street frontage extends to the first clearly visible perpendicular intersection. If none is visible, stop at the last visible point of the street-facing boundary. A pedestrian entrance is either: - A door in a building (including shop entrances) giving direct access from the street, or - A pedestrian gate in a fence or wall giving access from the street to an enclosed open area. Garage doors and car-only gates are NOT pedestrian entrances. Exception: if a car gate is the only access to a fenced plot and has a doorbell, street number, or mailbox, count it as one pedestrian entrance. If a pedestrian gate gives access to an enclosed open space before a building, count only the gate, not the building doors behind it. Pedestrian entrances partially hidden by foreground objects (trees, cars, people, street furniture) should still be considered if clear visual cues are present. \\
\midrule
FORMAT (std) & Answer with only one integer number, nothing else. \\
\midrule
FORMAT (reas) & (auto-generated by code --- do not fill in) \\
\bottomrule
\end{tabular}
\end{center}
}

\vspace{10pt}
\noindent\textbf{Task 4 --- Street Frontage Length  }
{\small
\begin{center}
\begin{tabular}{p{3cm}p{11cm}}
\midrule
Type / Column & task\_type: numeric    |    CSV column: length \\
\midrule
ROLE & (shared ROLE above) \\
\midrule
TASK & Estimate the total length (in meters) of the street frontage visible in the image. \\
\midrule
THEORY & The street frontage extends to the first clearly visible perpendicular intersection. If none is visible, stop at the last visible point of the street-facing boundary. The length is the sum of the linear lengths of all material and immaterial boundary elements along the street, from the left edge to the right edge of the visible street frontage. Only the street-facing boundary line is measured. Elements behind this boundary (such as building facades behind setbacks) are not included in the length, but may be used as visual references. Choose ONE type of standard-size reference object visible near the street-facing boundary and use it consistently to estimate the full length: - Cars (4 to 4.5 m each) - Parking bays (about 5 m each) - Doors or windows (about 1 m wide each) - Sidewalk slabs or modules - Building stories projected horizontally (about 3 m per story) - Electricity posts (about 8 m tall) or streetlamps (4 to 8 m tall), projected onto the horizontal boundary line. Account for perspective: objects farther from the camera appear shorter than their actual size. \\
\midrule
FORMAT (std) & Answer with only one integer number (meters), nothing else. \\
\midrule
FORMAT (reas) & (auto-generated by code --- do not fill in) \\
\bottomrule
\end{tabular}
\end{center}
}

\vspace{100pt}
\noindent\textbf{Task 5 --- Vegetation Classification}
{\small
\begin{center}
\begin{tabular}{p{3cm}p{11cm}}
\midrule
Type / Column & task\_type: numeric    |    CSV column: vegetation\_type \\
\midrule
ROLE & (shared ROLE above) \\
\midrule
TASK & Classify the vegetation presence along the street frontage visible in the image into one of six types. \\
\midrule
THEORY & The street frontage extends to the first clearly visible perpendicular intersection. If none is visible, stop at the last visible point of the street-facing boundary. First determine two things:  A. Trees in the street space: Trees are in the street space only if located in the street right-of-way along the frontage (on the sidewalk, curb strip, planting strip, or median — between the carriageway and the plot boundary). Trees inside plots (behind the plot boundary) do NOT count as trees in the street space. Only count trees large enough to form a canopy that at least partially covers the street space. If the image was taken in winter, deciduous trees may have no leaves. Still count them as trees if their trunk and branch structure indicate they are large enough to produce a canopy during the growing season.  B. Vegetation in plots: Vegetation in plots refers to trees, shrubs, lawn, or hedges seen behind the street-facing boundary inside the abutting plots. In winter, consider deciduous trees and shrubs as vegetation even if leafless.  Then select the matching type: Type 1 = Trees in street space YES + Plots highly vegetated (almost entirely covered) Type 2 = Trees in street space NO  + Plots highly vegetated Type 3 = Trees in street space YES + Plots partially vegetated (vegetated setbacks or side gardens) Type 4 = Trees in street space NO  + Plots partially vegetated Type 5 = Trees in street space YES + Plots little or no vegetation Type 6 = Trees in street space NO  + Plots little or no vegetation. \\
\midrule
FORMAT (std) & Answer with only one integer number (1 to 6), nothing else. \\
\midrule
FORMAT (reas) & (auto-generated by code --- do not fill in) \\
\bottomrule
\end{tabular}
\end{center}
}

\clearpage
\section*{Supplementary Material 2: Benchmark results and methodology}
\addcontentsline{toc}{section}{Supplementary Material 2}

\paragraph{Dataset} The benchmark dataset contains 120 street-level images of French urban frontages (IDs D01--D48, N01--N04, S01--S68)\cite{ref14}. Ground truth annotations were produced by a single expert annotator (Fusco.G). Some images were excluded from Qwen-3B runs due to CUDA crashes on a T4 GPU.
Due to computational cost constraints on an A100 GPU, only 25 images were randomly evaluated for LLaVA-34B reasoning mode.

\paragraph{Tasks} The benchmark evaluates the five visual interpretation tasks detailed in Supplementary Material 1.

\paragraph{Inference Modes} Two inference configurations are evaluated.

Standard mode. Direct answer generation. Maximum generation length: 50 tokens. Each task is executed three times, and the final prediction is obtained using majority vote (consensus).

Reasoning mode. Chain-of-thought reasoning followed by a final answer. Maximum generation length: 1024 tokens. The model is instructed to reason step-by-step and end with: ANSWER: <value> Three independent runs are executed and aggregated using majority vote.

\paragraph{Evaluation Metrics}

\textit{Binary classification (Sidewalk):} Accuracy: Fraction of images where the model prediction matches the human annotation. Sensitivity: True positive rate: proportion of sidewalk-present images correctly detected. Specificity: True negative rate: proportion of sidewalk-absent images correctly identified. Cohen's $\kappa$: Chance-corrected agreement between model and human annotations.

\textit{Continuous and counting tasks:} MAE (Mean Absolute Error): Average magnitude of prediction errors. Lower values indicate better performance. Bias: Mean signed error (model $-$ human). Positive values indicate systematic overestimation; negative values indicate underestimation. MAPE (Mean Absolute Percentage Error): MAE expressed as a percentage of the true value.
Particularly relevant for frontage length where absolute errors scale with size. Exact match: Fraction of images where the model prediction exactly matches the human value. $\pm$1 / $\pm$2 tolerance: Fraction of predictions within $\pm$1 or $\pm$2 units of the human value. $\pm$10 m tolerance: Equivalent tolerance metric for frontage length. Pearson correlation (r): Measures the linear association between model and human values.

\textit{Ordinal classification (Vegetation):} Weighted Cohen's $\kappa$ (linear weights): Measures agreement for ordered categories while penalizing large misclassifications more strongly than small ones. Interpretation follows the same scale as standard $\kappa$.

\textit{Proximity Score:} The main model ranking indicator is the Proximity Score, defined as: The range corresponds to the observed human value range for each task (e.g., 0--8 vehicles, 4--60 m frontage, 1--6 vegetation classes). Unlike Exact Match, this metric rewards near-correct predictions. The Overall Proximity Score is computed as the unweighted mean across the five tasks and is used as the primary model ranking criterion.

\textit{Reasoning Impact:} To evaluate the effect of chain-of-thought reasoning, we compute:
 NA count: Number of consensus runs where the parser fails to extract a valid answer. Truncation count: Number of task $\times$ image pairs where generation reached the maximum token limit. Truncation does not necessarily imply an incorrect answer, but it increases the probability of parsing failures because the reasoning chain may be incomplete.

\begin{table}[H]
\centering
\caption{Cost-performance analysis (*computing costs (CC) based on Google Colab GPU)}
\small
\begin{tabular}{llclccc}
\toprule
\textbf{Model} & \textbf{GPU} & \textbf{Load (min)} & \textbf{Mode} & \textbf{Sec/image} & \textbf{CC*/image} & \textbf{Proximity} \\
\midrule
Qwen 3B & T4 & 0.61 & Standard & \textcolor{green!60!black}{0.73} & 0.00033 & 76.0\% \\
Qwen 3B & T4 & 0.61 & Reasoning & 17.23 & 0.00644 & 76.1\% \\
Qwen 7B & T4 & 1.11 & Standard & 2.17 & 0.00090 & 75.4\% \\
Qwen 7B & T4 & 1.11 & Reasoning & 16.23 & 0.00779 & 77.4\% \\
Qwen 32B & A100 & 3.66 & Standard & 1.06 & 0.00519 & 83.3\% \\
Qwen 32B & A100 & 3.66 & Reasoning & 60.50 & 0.14031 & \textcolor{green!60!black}{88.0\%} \\
LLaVA Mistral 7B & T4 & 1.11 & Standard & 5.60 & 0.00269 & 74.4\% \\
LlaVA Mistral 7B & T4 & 1.11 & Reasoning & 26.31 & 0.01401 & 73.3\% \\
LlaVA Vicuna 7B & T4 & 1.07 & Standard & 3.03 & 0.00314 & 83.1\% \\
LlaVA Vicuna 7B & T4 & 1.07 & Reasoning & 17.99 & 0.04031 & 77.0\% \\
LlaMA3 LlaVA 8B & L4 & 1.20 & Standard & 1.62 & \textcolor{green!60!black}{0.00029} & 65.3\% \\
LlaMA3 LlaVA 8B & L4 & 1.20 & Reasoning & 27.16 & 0.01522 & 71.4\% \\
LlaVA Vicuna 13B & A100 & 1.63 & Standard & 1.19 & 0.00314 & 77.9\% \\
LlaVA Vicuna 13B & A100 & 1.63 & Reasoning & 19.30 & 0.04031 & 77.0\% \\
LlaVA 34B & A100 & 3.70 & Standard & 5.36 & 0.01191 & \textcolor{red!70!black}{62.2\%} \\
LLaVA 34B & A100 & 3.70 & Reasoning & \textcolor{red!70!black}{160.01} & \textcolor{red!70!black}{0.56122} & 74.2\% \\
\bottomrule
\end{tabular}
\end{table}

\begin{table}[H]
\centering
\caption{Task 1: Motor Vehicles}
\small
\begin{tabular}{lccccccc}
\toprule
\textbf{Model} & \textbf{N} & \textbf{MAE} & \textbf{Bias} & \textbf{Exact} & \textbf{$\pm$1} & \textbf{Correlation} & \textbf{Proximity} \\
\midrule
Qwen 3B std & 60 & 0.38 & -0.38 & \textcolor{green!60!black}{81.7\%} & 91.7\% & 0.90 & 96.5\% \\
Qwen 3B reas & 118 & 0.43 & -0.28 & 70.3\% & 92.4\% & 0.88 & 96.1\% \\
Qwen 7B std & 120 & 0.28 & -0.21 & 80.8\% & 95.8\% & 0.92 & 97.5\% \\
Qwen 7B reas & 120 & 0.47 & -0.43 & 70.8\% & 92.5\% & 0.87 & 95.8\% \\
Qwen 32B std & 120 & \textcolor{green!60!black}{0.21} & -0.11 & \textcolor{green!60!black}{81.7\%} & \textcolor{green!60!black}{98.3\%} & \textcolor{green!60!black}{0.96} & \textcolor{green!60!black}{98.1\%} \\
Qwen 32B reas & 120 & 0.25 & -0.20 & \textcolor{green!60!black}{81.7\%} & 95.0\% & 0.95 & 97.7\% \\
LLaVA Mis7B std & 120 & 0.47 & -0.40 & 68.3\% & 90.8\% & 0.88 & 95.8\% \\
LLaVA Mis7B reas & 119 & 0.66 & -0.51 & 57.1\% & 85.7\% & 0.82 & 94.0\% \\
LLaVA Vic7B std & 120 & 0.51 & -0.14 & 63.3\% & 92.5\% & 0.87 & 95.4\% \\
LLaVA Vic7B reas & 118 & 0.77 & -0.19 & \textcolor{red!70!black}{54.2\%} & \textcolor{red!70!black}{83.1\%} & 0.78 & 93.0\% \\
LLaMA3 8B std & 120 & 0.45 & -0.32 & 67.5\% & 92.5\% & 0.89 & 95.9\% \\
LLaMA3 8B reas & 120 & 0.69 & \textcolor{green!60!black}{-0.09} & 55.8\% & 86.7\% & 0.76 & 93.7\% \\
LLaVA Vic13B std & 120 & 0.43 & -0.23 & 73.3\% & 92.5\% & 0.85 & 96.1\% \\
LLaVA Vic13B reas & 120 & 0.71 & -0.44 & 58.3\% & 83.3\% & 0.77 & 93.6\% \\
LLaVA 34B std & 120 & 0.47 & -0.28 & 73.3\% & 91.7\% & 0.74 & 95.8\% \\
LLaVA 34B reas & 25 & \textcolor{red!70!black}{1.80} & \textcolor{red!70!black}{1.56} & 64.0\% & 88.0\% & \textcolor{red!70!black}{0.72} & \textcolor{red!70!black}{88.0\%} \\
\bottomrule
\end{tabular}
\end{table}

\begin{table}[H]
\centering
\caption{Task 2: Sidewalk}
\small
\begin{tabular}{lcccccc}
\toprule
\textbf{Model} & \textbf{N} & \textbf{Accuracy} & \textbf{Sensitivity} & \textbf{Specificity} & \textbf{Cohen $\kappa$} & \textbf{Proximity} \\
\midrule
Qwen 3B std & 60 & 75.0\% & 93.0\% & 29.4\% & \textcolor{green!60!black}{0.27} & 75.0\% \\
Qwen 3B reas & 118 & 56.8\% & 63.6\% & 21.1\% & \textcolor{red!70!black}{-0.11} & 56.8\% \\
Qwen 7B std & 120 & 40.0\% & 31.7\% & 84.2\% & 0.07 & 40.0\% \\
Qwen 7B reas & 120 & 62.5\% & 62.4\% & 63.2\% & 0.15 & 62.5\% \\
Qwen 32B std & 120 & 80.8\% & 90.1\% & 31.6\% & 0.23 & 80.8\% \\
Qwen 32B reas & 120 & \textcolor{green!60!black}{85.0\%} & 97.0\% & 21.1\% & 0.24 & \textcolor{green!60!black}{85.0\%} \\
LLaVA Mis7B std & 120 & 45.0\% & 42.6\% & 57.9\% & 0.00 & 45.0\% \\
LLaVA Mis7B reas & 120 & 70.0\% & 74.3\% & 47.4\% & 0.16 & 70.0\% \\
LLaVA Vic7B std & 120 & 84.2\% & \textcolor{green!60!black}{100.0\%} & \textcolor{red!70!black}{0.0\%} & 0.00 & 84.2\% \\
LLaVA Vic7B reas & 120 & 77.5\% & 85.1\% & 36.8\% & 0.21 & 77.5\% \\
LLaMA3 8B std & 120 & 51.7\% & 43.6\% & 94.7\% & 0.17 & 51.7\% \\
LLaMA3 8B reas & 120 & 46.7\% & 38.6\% & 89.5\% & 0.12 & 46.7\% \\
LLaVA Vic13B std & 120 & 80.8\% & 93.1\% & 15.8\% & 0.11 & 80.8\% \\
LLaVA Vic13B reas & 120 & 65.8\% & 68.3\% & 52.6\% & 0.14 & 65.8\% \\
LLaVA 34B std & 120 & \textcolor{red!70!black}{15.8\%} & \textcolor{red!70!black}{0.0\%} & \textcolor{green!60!black}{100.0\%} & 0.00 & \textcolor{red!70!black}{15.8\%} \\
LLaVA 34B reas & 25 & 80.0\% & 95.0\% & 20.0\% & 0.19 & 80.0\% \\
\bottomrule
\end{tabular}
\end{table}

\begin{table}[H]
\centering
\caption{Task 3: Pedestrian Entrances}
\small
\begin{tabular}{lccccccc}
\toprule
\textbf{Model} & \textbf{N} & \textbf{MAE} & \textbf{Bias} & \textbf{Exact} & \textbf{$\pm$1} & \textbf{Correlation} & \textbf{Proximity} \\
\midrule
Qwen 3B std & 60 & 1.35 & -1.28 & 26.7\% & 61.7\% & 0.67 & 77.5\% \\
Qwen 3B reas & 118 & 1.16 & -0.92 & 37.3\% & 72.9\% & 0.44 & 85.5\% \\
Qwen 7B std & 120 & 0.88 & -0.67 & 45.0\% & 79.2\% & 0.71 & 89.0\% \\
Qwen 7B reas & 120 & 1.24 & -0.59 & 34.2\% & 70.8\% & 0.56 & 84.9\% \\
Qwen 32B std & 120 & \textcolor{green!60!black}{0.62} & -0.28 & \textcolor{green!60!black}{50.0\%} & \textcolor{green!60!black}{90.8\%} & \textcolor{green!60!black}{0.84} & \textcolor{green!60!black}{92.3\%} \\
Qwen 32B reas & 120 & 0.77 & -0.42 & 49.2\% & 84.2\% & 0.71 & 90.4\% \\
LLaVA Mis7B std & 120 & 1.14 & -0.91 & 33.3\% & 74.2\% & 0.55 & 85.7\% \\
LLaVA Mis7B reas & 120 & 1.30 & -0.52 & 32.5\% & 69.2\% & 0.16 & 83.9\% \\
LLaVA Vic7B std & 120 & 1.14 & -0.29 & 31.7\% & 75.8\% & 0.27 & 85.7\% \\
LLaVA Vic7B reas & 120 & 1.53 & -0.44 & \textcolor{red!70!black}{22.5\%} & \textcolor{red!70!black}{52.5\%} & 0.32 & 80.9\% \\
LLaMA3 8B std & 120 & 1.12 & 0.43 & 26.7\% & 71.7\% & 0.59 & 86.0\% \\
LLaMA3 8B reas & 117 & 1.21 & 0.38 & 29.1\% & 71.8\% & 0.55 & 84.9\% \\
LLaVA Vic13B std & 120 & 1.10 & \textcolor{green!60!black}{-0.08} & 26.7\% & 77.5\% & 0.47 & 86.3\% \\
LLaVA Vic13B reas & 120 & 1.54 & 0.26 & 25.0\% & 57.5\% & 0.28 & 80.7\% \\
LLaVA 34B std & 120 & 1.13 & 0.62 & 23.3\% & 75.0\% & 0.60 & 85.8\% \\
LLaVA 34B reas & 25 & \textcolor{red!70!black}{2.64} & \textcolor{red!70!black}{1.60} & 36.0\% & 56.0\% & \textcolor{red!70!black}{0.15} & \textcolor{red!70!black}{76.5\%} \\
\bottomrule
\end{tabular}
\end{table}

\begin{table}[H]
\centering
\caption{Task 4: Street Frontage Length}
\small
\begin{tabular}{lccccccc}
\toprule
\textbf{Model} & \textbf{N} & \textbf{MAE (m)} & \textbf{Bias (m)} & \textbf{MAPE \%} & \textbf{$\pm$10m} & \textbf{Correlation} & \textbf{Proximity} \\
\midrule
Qwen 3B std & 60 & 11.4 & 6.1 & 81.8 & 73.3\% & \textcolor{red!70!black}{-0.14} & 70.1\% \\
Qwen 3B reas & 118 & 13.0 & 5.2 & 81.5 & 61.9\% & 0.08 & 66.5\% \\
Qwen 7B std & 120 & 10.5 & 7.7 & 63.8 & 74.2\% & 0.41 & 70.1\% \\
Qwen 7B reas & 120 & 9.0 & -6.2 & 47.3 & 69.2\% & 0.29 & 73.3\% \\
Qwen 32B std & 120 & 16.5 & 16.0 & 106.1 & 50.0\% & \textcolor{green!60!black}{0.49} & 55.4\% \\
Qwen 32B reas & 120 & \textcolor{green!60!black}{5.8} & \textcolor{green!60!black}{0.1} & \textcolor{green!60!black}{35.8} & \textcolor{green!60!black}{85.0\%} & 0.48 & \textcolor{green!60!black}{82.5\%} \\
LLaVA Mis7B std & 120 & 10.9 & 0.9 & 53.7 & 74.2\% & 0.30 & 75.4\% \\
LLaVA Mis7B reas & 120 & 12.5 & -5.5 & 65.9 & 52.5\% & 0.13 & 65.9\% \\
LLaVA Vic7B std & 120 & 9.5 & -2.1 & 48.1 & 76.7\% & 0.06 & 76.5\% \\
LLaVA Vic7B reas & 120 & \textcolor{red!70!black}{53.2} & \textcolor{red!70!black}{37.8} & \textcolor{red!70!black}{333.8} & 55.0\% & -0.02 & 67.1\% \\
LLaMA3 8B std & 120 & 42.9 & 41.8 & 258.4 & 24.2\% & 0.41 & \textcolor{red!70!black}{32.8\%} \\
LLaMA3 8B reas & 120 & 11.7 & -6.6 & 66.7 & 53.3\% & 0.07 & 66.3\% \\
LLaVA Vic13B std & 116 & 27.7 & 21.3 & 150.1 & 56.0\% & 0.36 & 56.0\% \\
LLaVA Vic13B reas & 120 & 10.8 & -5.1 & 58.7 & 65.0\% & 0.13 & 70.4\% \\
LLaVA 34B std & 120 & 17.7 & -17.7 & 98.3 & \textcolor{red!70!black}{16.7\%} & -0.03 & 47.2\% \\
LLaVA 34B reas & 25 & 10.6 & 2.9 & 65.8 & 72.0\% & 0.22 & 70.4\% \\
\bottomrule
\end{tabular}
\end{table}

\begin{table}[H]
\centering
\caption{Task 5: Vegetation Type}
\small
\begin{tabular}{lcccccc}
\toprule
\textbf{Model} & \textbf{N} & \textbf{Exact} & \textbf{$\pm$1 class} & \textbf{MAE class} & \textbf{$\kappa$ linear} & \textbf{Proximity} \\
\midrule
Qwen 3B std & 60 & 38.3\% & 43.3\% & 1.97 & 0.26 & 60.7\% \\
Qwen 3B reas & 118 & 37.3\% & 57.6\% & 1.23 & 0.31 & 75.4\% \\
Qwen 7B std & 120 & 47.5\% & 63.3\% & 0.97 & 0.42 & 80.7\% \\
Qwen 7B reas & 120 & 26.7\% & 36.7\% & 1.47 & 0.18 & 70.7\% \\
Qwen 32B std & 120 & \textcolor{green!60!black}{68.3\%} & \textcolor{green!60!black}{80.8\%} & \textcolor{green!60!black}{0.51} & \textcolor{green!60!black}{0.64} & \textcolor{green!60!black}{89.8\%} \\
Qwen 32B reas & 120 & 55.8\% & 66.7\% & 0.79 & 0.54 & 84.2\% \\
LLaVA Mis7B std & 120 & 27.5\% & 55.0\% & 1.48 & 0.25 & 70.3\% \\
LLaVA Mis7B reas & 120 & 13.3\% & \textcolor{red!70!black}{29.2\%} & \textcolor{red!70!black}{2.36} & 0.00 & \textcolor{red!70!black}{52.8\%} \\
LLaVA Vic7B std & 120 & 15.8\% & 73.3\% & 1.30 & \textcolor{red!70!black}{-0.02} & 74.0\% \\
LLaVA Vic7B reas & 120 & 27.5\% & 47.5\% & 1.68 & 0.10 & 66.3\% \\
LLaMA3 8B std & 120 & 16.7\% & 42.5\% & 1.98 & 0.00 & 60.3\% \\
LLaMA3 8B reas & 120 & 15.0\% & 48.3\% & 1.77 & 0.19 & 65.3\% \\
LLaVA Vic13B std & 120 & 20.0\% & 60.8\% & 1.48 & 0.05 & 70.3\% \\
LLaVA Vic13B reas & 120 & 30.8\% & 60.8\% & 1.28 & 0.21 & 74.5\% \\
LLaVA 34B std & 120 & \textcolor{red!70!black}{10.8\%} & 54.2\% & 1.68 & 0.05 & 66.5\% \\
LLaVA 34B reas & 25 & 16.0\% & 40.0\% & \textcolor{red!70!black}{2.36} & 0.06 & 56.0\% \\
\bottomrule
\end{tabular}
\end{table}

\begin{table}[H]
\centering
\caption{Reasoning vs Standard: Proximity delta per task}
\small

\resizebox{\textwidth}{!}{
\begin{tabular}{lccccccc}
\toprule
\textbf{Model pair} & \textbf{Vehicles} & \textbf{Sidewalk} & \textbf{Entrances} & \textbf{Length} & \textbf{Vegetation} & \textbf{Mean $\Delta$} & \textbf{Verdict} \\
\midrule
Qwen 3B std $\to$ Qwen 3B reas & -0.4\% & -18.2\% & +8.0\% & -3.6\% & +14.8\% & +0.1\% & Neutral \\
Qwen 7B std $\to$ Qwen 7B reas & -1.7\% & +22.5\% & -4.1\% & +3.2\% & -10.0\% & +2.0\% & \textcolor{green!60!black}{Helps} \\
Qwen 32B std $\to$ Qwen 32B reas & -0.4\% & +4.2\% & -1.9\% & +27.1\% & -5.7\% & +4.7\% & \textcolor{green!60!black}{Helps} \\
LLaVA Mis7B std $\to$ LLaVA Mis7B reas & -1.8\% & +25.0\% & -1.9\% & -9.5\% & -17.5\% & -1.1\% & \textcolor{red!70!black}{Hurts} \\
LLaVA Vic7B std $\to$ LLaVA Vic7B reas & -2.4\% & -6.7\% & -4.8\% & -9.4\% & -7.7\% & -6.2\% & \textcolor{red!70!black}{Hurts} \\
LLaMA3 8B std $\to$ LLaMA3 8B reas & -2.2\% & -5.0\% & -1.1\% & +33.5\% & +5.0\% & +6.0\% & \textcolor{green!60!black}{Helps} \\
LLaVA Vic13B std $\to$ LLaVA Vic13B reas & -2.6\% & -15.0\% & -5.5\% & +14.4\% & +4.2\% & -0.9\% & Neutral \\
LLaVA 34B std $\to$ LLaVA 34B reas & -7.8\% & +64.2\% & -9.3\% & +23.1\% & -10.5\% & +11.9\% & \textcolor{green!60!black}{Helps} \\
\bottomrule
\end{tabular}
}
\end{table}

\begin{table}[H]
\centering
\caption{Parsing failures (NA counts across all consensus runs)}
\small
\resizebox{\textwidth}{!}{
\begin{tabular}{lcccccccc}
\toprule
\textbf{Model} & \textbf{N images} & \textbf{Vehicles} & \textbf{Sidewalk} & \textbf{Length} & \textbf{Entrances} & \textbf{Vegetation} & \textbf{Total NAs} & \textbf{NA rate} \\
\midrule
Qwen 3B std & 60 & 0 & 0 & 0 & 0 & 0 & 0 & 0.0\% \\
Qwen 3B reas & 118 & 11 & 0 & 1 & 2 & 0 & 14 & 0.8\% \\
Qwen 7B std & 120 & 0 & 0 & 0 & 0 & 0 & 0 & 0.0\% \\
Qwen 7B reas & 120 & 0 & 0 & 0 & 0 & 0 & 0 & 0.0\% \\
Qwen 32B std & 120 & 0 & 0 & 0 & 0 & 0 & 0 & 0.0\% \\
Qwen 32B reas & 120 & 0 & 0 & 0 & 0 & 0 & 0 & 0.0\% \\
LLaVA Mis7B std & 120 & 0 & 0 & 0 & 0 & 0 & 0 & 0.0\% \\
LLaVA Mis7B reas & 120 & 9 & 0 & 8 & 26 & 0 & 43 & 2.4\% \\
LLaVA Vic7B std & 120 & 0 & 0 & 0 & 0 & 0 & 0 & 0.0\% \\
LLaVA Vic7B reas & 120 & 87 & 0 & 9 & 13 & 1 & 110 & 6.1\% \\
LLaMA3 8B std & 120 & 0 & 0 & 5 & 0 & 0 & 5 & 0.3\% \\
LLaMA3 8B reas & 120 & 68 & 0 & 0 & 87 & 41 & 196 & 10.9\% \\
LLaVA Vic13B std & 120 & 0 & 0 & 80 & 0 & 0 & 80 & 4.4\% \\
LLaVA Vic13B reas & 120 & 18 & 0 & 11 & 13 & 0 & 42 & 2.3\% \\
LLaVA 34B std & 120 & 0 & 0 & 0 & 1 & 0 & 1 & 0.1\% \\
LLaVA 34B reas & 25 & 0 & 0 & 3 & 1 & 1 & 5 & 1.3\% \\
\bottomrule
\end{tabular}
}
\end{table}

\begin{table}[H]
\centering
\caption{Truncation alerts}
\small

\resizebox{\textwidth}{!}{
\begin{tabular}{lcccccccc}
\toprule
\textbf{Model} & \textbf{N images} & \textbf{Vehicles} & \textbf{Sidewalk} & \textbf{Length} & \textbf{Entrances} & \textbf{Vegetation} & \textbf{Total} & \textbf{Trunc. rate} \\
\midrule
Qwen 3B std & 60 & 0 & 0 & 0 & 0 & 0 & 0 & 0.0\% \\
Qwen 3B reas & 118 & 1 & 0 & 0 & 1 & 0 & 2 & 0.3\% \\
Qwen 7B std & 120 & 0 & 0 & 0 & 0 & 0 & 0 & 0.0\% \\
Qwen 7B reas & 120 & 0 & 0 & 3 & 0 & 0 & 3 & 0.5\% \\
Qwen 32B std & 120 & 0 & 0 & 0 & 0 & 0 & 0 & 0.0\% \\
Qwen 32B reas & 120 & 0 & 0 & 0 & 0 & 0 & 0 & 0.0\% \\
LLaVA Mis7B std & 120 & 0 & 0 & 0 & 0 & 0 & 0 & 0.0\% \\
LLaVA Mis7B reas & 120 & 0 & 0 & 10 & 9 & 0 & 19 & 3.2\% \\
LLaVA Vic7B std & 120 & 0 & 0 & 0 & 0 & 0 & 0 & 0.0\% \\
LLaVA Vic7B reas & 120 & 0 & 0 & 4 & 1 & 0 & 5 & 0.8\% \\
LLaMA3 8B std & 120 & 0 & 0 & 3 & 0 & 0 & 3 & 0.5\% \\
LLaMA3 8B reas & 120 & 9 & 30 & 11 & 31 & 20 & 101 & 16.8\% \\
LLaVA Vic13B std & 120 & 0 & 0 & 21 & 0 & 0 & 21 & 3.5\% \\
LLaVA Vic13B reas & 120 & 0 & 0 & 6 & 1 & 1 & 8 & 1.3\% \\
LLaVA 34B std & 120 & 0 & 0 & 3 & 1 & 0 & 4 & 0.7\% \\
LLaVA 34B reas & 25 & 0 & 1 & 3 & 2 & 0 & 6 & 4.8\% \\
\bottomrule
\end{tabular}
}
\end{table}
\end{document}